\documentclass{article}
\usepackage[utf8]{inputenc}
\usepackage{nips_adapted}
\usepackage{pmisubmit}
\usepackage{pmi}
\usepackage[english]{babel}
\usepackage{bbm}
\usepackage{array}
\usepackage{float}
\usepackage{biblatex}
\usepackage{tabularx}
\usepackage{caption}
\usepackage{graphicx}
\usepackage{booktabs}
\usepackage{blindtext}

\nipsfinalcopy

\title{Self-Labeling Refinement for Robust Representation Learning with Bootstrap Your Own Latent}
\author{
Siddhant Garg \\
siddhantgarg@umass.edu \\
\And 
Dhruval Jain\\
dhruvaljain@umass.edu
}
\date{September 2021}
\begin{document}
\maketitle
\begin{abstract}
In this work, we have worked towards two major goals. Firstly, we have investigated the importance of Batch Normalisation (BN) layers in a non-contrastive representation learning framework called Bootstrap Your Own Latent (BYOL)\cite{byol}. We conducted several experiments to conclude that BN layers are not necessary for representation learning in BYOL \cite{NoStatistics}. Moreover, BYOL only learns from the positive pairs of images but ignore other semantically similar images in same input batch. For the second goal, we have introduced two new loss functions to determine the semantically similar pairs in the same input batch of images and reduce the distance between their representations. These loss functions are \textbf{Cross-Cosine Similarity Loss (CCSL)} and \textbf{Cross-Sigmoid Similarity Loss (CSSL)}. Using the proposed loss functions, we are able to surpass the performance of Vanilla BYOL (\textbf{71.04\%}) by training BYOL framework using CCSL loss (\textbf{76.87\%}) on \textbf{STL10} dataset. BYOL trained using CSSL loss performs comparably with Vanilla BYOL.

% Bootstrap Your Own Latent (BYOL) is a non-contrastive Self-Supervised Learning method that outperforms several state-of-the-art contrastive methods like SimCLR and Momentum Contrast (MoCo). BYOL uses a predicted network at the end of the online network and a slow-moving target for learning representations and avoid representation collapse. Some previous works suggested that the Batch Normalization(BN) layers are necessary for representation learning in BYOL and we conducted several experiments to prove  Furthermore, BYOL only learns from the positive pairs of images but ignore other semantically similar images in same input batch. We introduce two new loss functions to determine the semantically similar pairs in the same input batch of images and reduce the distance between their representations. Our linear evaluation protocol results show that we get higher validation accuracies on one of the proposed loss function and comparable accuracies on the second loss function. 
\end{abstract}
\section{Introduction}
Self-Supervised learning methods aim to leverage information from unlabeled dataset and improve the representation of the embeddings or hidden layers by learning semantic information about the inputs. Self-Supervised Learning methods have become quite popular in the domain of Computer Vision and many state-of-the-art methods, like SEER \cite{seer}, SwAV \cite{swav}, Contrastive Learning \cite{contrastivelearning} methods like SimCLR \cite{simclr} and Momentum Contrast (MoCo) \cite{he2020moco}, and Non-Constrastive Learning methods like BYOL\cite{byol} and RAFT\cite{raft} have surfaced over past years that are outperforming the traditional supervised learning methods on downstream tasks like Image Classification, Object Detection and Semantic Segmentation. 

Training deep neural networks with a Self-Supervised framework for Computer Vision tasks require us to design various pretext tasks like Context Prediction \cite{contextprediction}, or ordering of image grids \cite{unsupervisedjigsaw} to learn various concepts. The representation or features of similar concepts should be closer while the features of semantically different concepts should be further apart. This is achieved using $InfoNCE$ Loss function \cite{contrastivepredictive} and large number of negative samples for a given input image. However, a particular Contrastive Learning Method called Momentum Contrast (MoCo) \cite{he2020moco}, keeps memory bank of negative samples from the past mini-batches that allows it to reduce the batch size during training, and thus making it computationally feasible. Recenlty, a non-contrastive method called Bootstrap Your Own Latent(BYOL)\cite{byol}, was proposed that does not require negative samples at all for learning features. The method is also robust to the image augmentations and batch size while training.

The negative pairs in contrastive learning were necessary to avoid a representation collapse in which case the network outputs a trivial representation for all the input images. However, the absence of the negative pairs in BYOL stirred a lot of commotion in the community and several works \cite{NoStatistics, bn_1, tian2021understanding, raft} have tried to understand the phenomenon. It has been shown that BYOL avoids representation collapse by using a predictor network at the end of the online network and by updating its target network using slow-moving averages and not by backpropagating the gradients through the target network. We have also conducted several experiments to rule out the importance or necessity of Batch Normalization \cite{batchnorm} layers for the representation learning in BYOL, as suggested by \cite{bn_1}, by replacing them with Layer Normalization \cite{ln} and Group Normalization layers \cite{gn}.  

Furthermore, the MoCo framework assumes that if the key and query pair (images after applying data augmentations) are from the same image, then they are treated as positive pairs and rest of the keys, in the mini-batch, are treated as negative pair with the query. But, more than one images in a mini-batch can be semantically similar and by treating them as negative pairs, the quality of the representations are impaired. Zhou et al. \cite{selflabelrefinement} proposed a Self-Labeling Refinement Method to improve the label quality and learning more generalized representations. We will extend this method for BYOL by proposing two new loss functions and compare the results of the vanilla training method with the proposed Self-Label Refinement method \footnote{Work done for neural networks course project at UMass Amherst}
. 

\section{Related Work} \label{sec:related_work}
Learning unsupervised representations can be seen as learning a energy function or a contrast function that have low energy values on the data distribution and high energies everywhere else. Probabilistic models can be used to model such functions using Gibbs Distribution \cite{gibbs_dist} which is proportional to the exponential of the negative energy values. The normalization constant, however, is intractable for high dimensional data and they are approximated using MCMC methods \cite{gibbs_samp, mh, mcmc}. Sparse Autoencoders \cite{sparse_enc}, and Latent Variable Models \cite{latentvarmodels} are some of the examples that learn the latent representations of the unlabeled data by minimizing some form of energy function using encoder-decoder frameworks.

For Self-Supervised learning, several contrastive methods \cite{he2020moco, simclr, contrastivepredictivecoding, multiviewcoding, Dataefficientimagerecognition, mutualinformation, mutualinformationacrossviews, pretext-invariant, Prototypicalcontrastive} have been proposed that achieved state-of-the-art results. Contrastive learning methods bring the representations of the augmented views (positive pair) of the same image closer and augmented views of different images (negative pairs) further apart. SimCLR \cite{simclr} and MoCo \cite{he2020moco} are the most popular contrastive learning methods that use a loss function for contrastive prediction task called InfoNCE \cite{contrastivepredictive} loss which is defined using equation \ref{eq:nt-xent}. 

Self-Supervised learning can also be done by introducing some pretext tasks and perform predictions using those auxiliary tasks. Auxiliary tasks can include image jigsaw puzzle \cite{unsupervisedjigsaw}, in which the image is divided into grids and the task is to predict the correct ordering of the grids, colorization of grayscale images \cite{Colorfulimagecolorization, automaticcolorization}, image inpainting \cite{inpainting}, relative patch prediction \cite{patch1, patch2}, geometric transformations \cite{geomtransform1, geomtransform2} like image rotations, image super resolution \cite{super-resolution} or by using suitable architectures \cite{suitablearchitecture} like Prototypical networks\cite{Prototypicalcontrastive}. Though we can design variety of pretext tasks, the contrastive methods are the state-of-the-art for Self-Supervised learning. 

Though contrastive learning methods have been proven to achieve state-of-the-art results for the Self-Supervised learning but they require high computational resources. For a single positive pair, a large amount of negative samples are required with high batch size. For example, in SimCLR \cite{simclr}, for a single positive pair the batch size went up to 8192 and the model was trained upto 128 TPU cores. Momentum Contrast (MoCo) \cite{he2020moco} tries to alleviate this issue by using a dictionary of negative samples. However, Bootstrap Your Own Latent (BYOL) \cite{byol}, addresses the issue of high computational costs by removing the negative pairs for learning image representations. Instead BYOL uses a slow-moving average called target network to learn the image representations. Since, BYOL does not use negative pairs, an extra predictor head is used in the online network encoder to avoid representation collapse. BYOL outperformed several contrastive learning methods including SimCLR and MoCo on ImageNet \cite{deng2009imagenet} dataset and it has been shown to be robust to the batch size. MoCo also uses a slow moving target to update the key encoder which is necessary for learning representations. \textit{Mean Teacher(MT)} \cite{meanteacher} networks, that are used in semi-supervised learning, also use slow-moving averages as the target network called \textit{teacher} to train the online network called \textit{student}.  

There have been several attempts at analysing BYOL to understand its dynamics and to answer questions like why it does not collapse without the contrastive negative pairs. Tian et al \cite{tian2021understanding} performs several experiments and report that the predictor network at the end of the online network and stop-gradients for the target network are necessary to avoid the representation collapse. Some works \cite{bn_1} also suggest that the Batch Normalization layer is responsible for \textit{implicit contrastive learning} but experiments from \cite{NoStatistics} shows that Batch Normalization \cite{batchnorm} layers are not necessary to avoid the representation collapse. We conduct several experiments in this work by replacing the Batch Normalization layers \cite{batchnorm} with Layer Normalization \cite{ln} and Group Normalization \cite{gn} along with weight standardization \cite{weightstandardization}. 

Contrastive Learning methods learn the representations from the unlabeled data by considering the views of the same image as positive while taking all the other views as negative in the input batch of images. This can introduce some errors because some negative pairs in the same batch might be semantically similar to the positive pairs. Zhou et al \cite{selflabelrefinement} proposed two modules called Self-label refinery to generate accurate labels and Momentum Mixup to enhance the similarity between an image and its positive. These two modules were designed for MoCo. In this work, we proposed a novel Self-Labeling Refinement for BYOL by introducing two loss functions that identifies the semantically similar images in an input batch and minimize the representation loss between them. 

\section{Technical Approach and Methods} \label{sec:methods}
In this section, we will describe the details of BYOL network and details of the proposed methods.
% For each of the methods, we will also provide the details of the Self-Labeling Refinement to learn more robust representations. For MoCo, we will follow, the method proposed by Zhuo et al \cite{selflabelrefinement} and for BYOL, a new loss function is proposed in this paper, for implicitly incorporating semantic similarity and dissimilarity in image in the input mini-batch.

\subsection{Bootstrap Your Own Latent(BYOL)}
\label{sec:byol}
\begin{figure}[t]
    \centering
\includegraphics[width=0.7\linewidth]{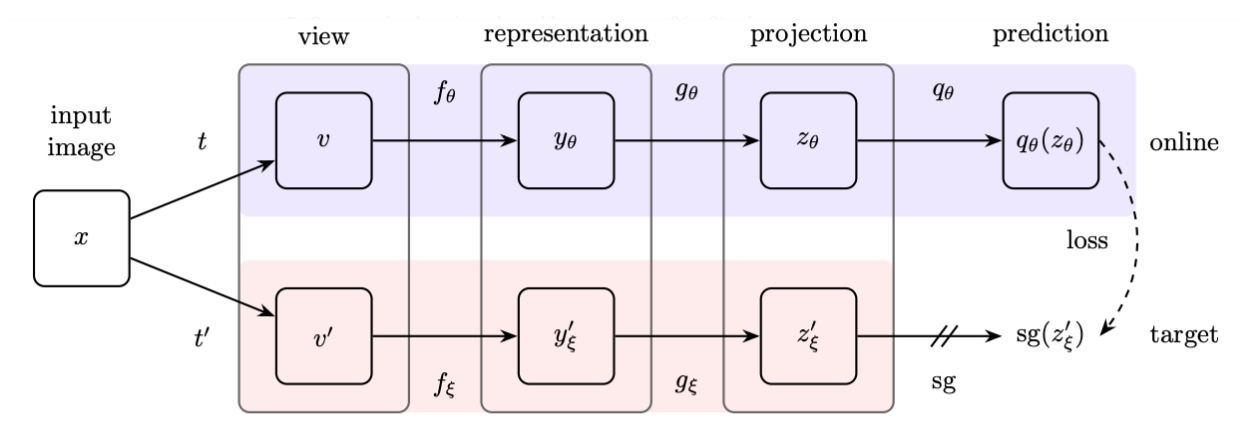}
    \caption{The figure (from \cite{byol}) represents the framework of BYOL. Two views of the same image are passed through online and target networks and their is predictor at the end of online network, which prevents collapsed representations. $sg$ stands for $stop-gradient$ because the gradients will get backpropagated through the online network only, whereas the target network will be updated using the slow-moving average of the online network. The loss function minimizes the $L2$ norm between the outputs of the two networks.}
    \label{fig:byol_acrch}
\end{figure}
\subsubsection{Representation Learning}
\label{sec:representation_learning}
BYOL \cite{byol} uses two neural networks called online network, represented by parameters $\theta$ and target network, represented by $\xi$. The online network consists of an encoder $f_{\theta}$, projector $g_{\theta}$ and a predictor $q_{\theta}$ and target network consists of and encoder $f_{\xi}$ and a projector $g(\xi)$. The representations from the target network are used to obtain more enhanced representations by predicting the target representations using the online networks. The weights of the online network are back-propagated and the target network is updated using the slow moving average of the online network with the decay rate of $\tau \in [0,1]$ using equation \ref{eq:opt_xi}.

% Online network is represented by $\theta$ and the target network is represented by $\xi$. $f_{\theta}$ and $f_{\xi}$ denote online and target encoders respectively. The target network weights are calculated by moving average weights of the online network using the equation \ref{eq:movingavg}. 
% \begin{align} \label{eq:movingavg}
%     \xi = \tau\xi + (1-\tau)\theta
% \end{align}

Given a set of input images $\mathcal{D}$ and two image augmentations $\mathcal{T}$ and $\mathcal{T'}$, two different views of the same image, called $v = t(x)$ and $v'=t'(x)$ can be generated using $t \sim T$ and $t'\sim T'$. Using the two augmented views, the online network produces an output $q_{\theta}(z_{\theta})$ and the target network produces the output $z'_{\xi}$. Both the outputs are normalized (getting $\Bar{q_{\theta}}(z_{\theta})$ and $\Bar{z'}_{\xi}$) and the loss $\mathcal{L_{\theta, \xi}}$ is calculated by minimizing the $L2$ loss between the normalized outputs. 
\begin{align}
    \mathcal{L_{\theta, \xi}} &= \norm{\Bar{q_{\theta}}(z_{\theta}) - \Bar{z'}_{\xi}}_2 = 2 - 2.\frac{\langle q_{\theta}(z_{\theta}), z'_{\xi}\rangle}{\norm{q_{\theta}(z_{\theta})}_2. \norm{z'_{\xi}}_2}
\end{align}
Note that we only have a predictor $q_{\theta}$ in the online network therefore, the architecture is asymmetric between the online network and the target network. To symmetrize the loss, the inputs, $v$ and $v'$ are swapped and $v$ is fed to the target network and $v'$ is fed to the online network to get the loss $\tilde{\mathcal{L}}_{\theta, \xi}$. Combining both the losses we get the final loss function, $\mathcal{L}^{byol}$.
\begin{align} \label{eq:byol_loss}
    \mathcal{L}^{byol}_{\theta, \xi} &= \mathcal{L}_{\theta, \xi} + \tilde{\mathcal{L}}_{\theta, \xi}
\end{align}
The gradients of the loss function in equation \ref{eq:byol_loss} are calculated with respect to $\theta$ only and not $\xi$. The complete optimization process can be summarized in equations \ref{eq:opt_theta} and \ref{eq:opt_xi}.  
\begin{align} 
    \theta &\leftarrow \text{optimizer}(\theta, \nabla_{\theta}\mathcal{L}^{byol}_{\theta, \xi}, \eta) \label{eq:opt_theta}\\
    \xi &\leftarrow \tau \xi + (1-\tau)\theta \label{eq:opt_xi}
\end{align}
% We will implement the BYOL network in PyTorch with ResNet-18 as our encoder module. Existing data augmentations proposed in the original paper will be used for preliminary experiments. 

\subsection{Batch Statistics} \label{sec:batch_statistics}
Contrastive learning focuses on minimizing the distance between intra-class images (positive pairs) while maximizing that distance metric for inter-class images (negative pairs). This paradigm has much gathered a lot of  attention in representation learning. However, BYOL framework does not use any contrastive term explicitly in its loss function for negative pairs. This is well illustrated from Eq (2). It is astonishing that still it is able to avoid learning a collapse solution, where the networks predicts the same constant representation for all samples. It has been hypothesised that Batch Normalisation \cite{bn} (BN) provides an implicit contrastive(negative) term. Inspired from \cite{NoStatistics}, we perform an ablation study by checking the performance of BYOL after removing all BN layers. Additionally, we have also applied Weight standardization (WS) \cite{ws} after convolution layers that were followed by BN layers in both the online and target networks. All the experiments are described in detail in section 4. 
\subsubsection{Weight Standardization (WS)}
The weight matrix $W$ from any convolution layer is normalised along the rows to produce $\widehat{W}$ which is then used for training. However, during the back-propogation, the gradient of the loss is computed only with respect to the non-normalized weights $W$.

\begin{align} \label{eq:WS}
    \widehat{W}_{i,j} &= \frac{W - \mu_{i,j}}{\sigma_{i,j}}, \hspace{2mm} \text{where} \hspace{1mm}\mu_{i} = \frac{1}{R} \sum_{j=1}^{R} W_{i,j} \hspace{2mm} \text{and} \hspace{2mm} \sigma_{i} = \sqrt{\epsilon + \frac{1}{R}\sum_{j=1}^{R} (W_{i,j} - \mu_{i})^2}
\end{align}

where $R$ is the input dimension which is equal to the product of input channel dimension and the filter size of the convolution layer. $\epsilon$ is typically of the order of $10^{-4}$. WS was originally proposed for applying BN with micro-batch training settings, but it also smoothens the loss landscape. In their paper, \cite{ws} have proved that WS reduces the Lipschitz constants of the loss and the gradients which results in faster convergence and reduction in training time. There are no learning parameters in WS unlike BN.

\subsection{Self-Labeling Refinement} \label{sec:self_labeling}
BYOL, does not  rely on negative samples to learn robust image representations but only minimizes the representational loss between the two different views of the same image. But consider the case in which there are semantically similar images (images containing dogs, or images containing only cats) in the same input mini-batch. Can we somehow identify those and make their representations closer? And whether doing this would improve the quality of the representations or would it lead to representation collapse? In this section, we propose two methods to identify the semantically similar images from the learned representations of the BYOL and further train the network using different loss functions. To identify the similar images we calculated the dot product of the normalized representations of every image with every other image using the representations of the online network and the target network. Since, the target network is just the moving exponential average of the online network we would expect that the similar images have higher dot product and dissimilar images have small values of the dot products. We defined thresholds $\theta_p$ and $\theta_n$ to keep most similar and dissimilar images. If the dot product is higher than $\theta_p$, then the images are similar, therefore, their representations should be closer and if the dot product is lower than $\theta_n$, then the images are dissimilar and there representations should be further apart. 

Now we will describe the two proposed loss functions below. First loss function just extends the $L2$ loss function to include the $L2$ loss between the similar and dissimilar images. Our second loss function extends the infoNCE \cite{contrastiveloss} loss to calculate the similarity and dissimilarity between the input images.  

\subsubsection{Cross-Cosine Similarity Loss}
\label{sec:ccsl}
The \textit{Cross-Cosine Similarity Loss(CCSL)} calculates the cosine loss between the output representations of the online and target network of BYOL which was trained earlier. For an input image $x_i$, the expressions of the loss function is described in in Eq \ref{eq:CCSL-complete}
\begin{equation}
\label{eq:CCSL-complete}
\begin{split}
    \mathcal{L}_{CCSL}(x_i) &= \norm{\Bar{q}_{z_\theta}(x_i) - \Bar{z}'_{\xi}(x_i)}_2 + \lambda\sum_{j\neq i} \mathbbm{1}[\langle \Bar{q}_{z_{\theta}}(x_i), \Bar{z}'_{\xi}(x_j)\rangle \geq \theta_p] \norm{\Bar{q}_{z_{\theta}}(x_i) - \Bar{z}'_{\xi}(x_j)}_2 \\
    & \hspace{60mm} - \lambda\sum_{j\neq i} \mathbbm{1}[\langle \Bar{q}_{z_{\theta}}(x_i), \Bar{z}'_{\xi}(x_j)\rangle \leq \theta_n] \norm{\Bar{q}_{z_{\theta}}(x_i) - \Bar{z}'_{\xi}(x_j)}_2
\end{split}
\end{equation}
Note that the Eq \ref{eq:CCSL-complete} consists of three terms. The first term is the standard $L2$ loss between the online and target representations of the same image. The second term is the cosine loss/distance between the highly similar images identified by the indicator function that checks the dot product between the $l2$-normalized outputs and if the distance is above a certain threshold $\theta_p$, we try to bring the representations even closer by adding the $L2$ loss between them. Similarly, when the cosine similarity between the representations is lower than a certain threshold $\theta_n$, we try to separate them out even more by adding the negative $L2$ loss between them. However, adding the negative $L2$ loss between the representations led to the representational collapse therefore, we removed this term and kept only the first two terms. Therefore, the final loss function is calculated using equation \ref{eq:CCSL-final}
\begin{equation}
\label{eq:CCSL-final}
\begin{split}
    \mathcal{L}_{CCSL}(x_i) &= \norm{\Bar{q}_{z_\theta}(x_i) - \Bar{z}'_{\xi}(x_i)}_2 + \lambda\sum_{j\neq i} \mathbbm{1}[\langle \Bar{q}_{z_{\theta}}(x_i), \Bar{z}'_{\xi}(x_j)\rangle \geq \theta_p] \norm{\Bar{q}_{z_{\theta}}(x_i) - \Bar{z}'_{\xi}(x_j)}_2 
\end{split}
\end{equation}
Note that we are weighted the second term of the loss with $\lambda$ which is set to a small value to avoid the representation collapse otherwise without it all the representations became equal which could also lead to a minimum loss. We are calling this loss function as \textit{Cross-Cosine Similarity Loss} because we are calculating the loss between $\mathcal{O}(N^2)$ terms, $N$ being the mini-batch size. Finally, to make the loss function symmetric we will also interchange the input values of the online network and target network and compute the CCSL loss just like we did in the vanilla BYOL setup. Let this value be called as $\tilde{\mathcal{L}}_{CCSL}$, then the final CCSL loss can written as in Eq \ref{eq:CCSL-both-term} 
\begin{equation}
\label{eq:CCSL-both-term}
    \mathcal{L}_{CCSL}^{BYOL}(x_i) = \mathcal{L}_{CCSL}(x_i) + \tilde{\mathcal{L}}_{CCSL}(x_i) 
\end{equation}
The loss over the mini-batch can be computed as the average of the loss in Eq \ref{eq:CCSL-both-term}

\subsubsection{Cross-Sigmoid Similarity Loss}
\label{sec:cssl}
In popular contrastive representation learning methods like SimCLR \cite{simclr}, the representation loss is calculated using the $NT-Xent$ (Normalized temperature scaled cross-entropy loss Eq \ref{eq:nt-xent}
\begin{equation}
\label{eq:nt-xent}
    \ell(x_i, \tilde{x}_j) = -\log \frac{exp (sim (z_i, \tilde{z}_j)/\tau)}{\sum_{k=1}^{2N}\mathbbm{1}[k\neq i]exp (sim (z_i, \tilde{z}_k)/\tau)}
\end{equation}
The $sim(.)$ in eq \ref{eq:nt-xent} is the cosine similarity, $\tau$ is the temperature parameter and the loss is calculated such that the two views of the same image should become similar and the views of the different images become different in the terms of the cosine similarity function. Note that the Eq \ref{eq:nt-xent} is similar to the popular cross entropy loss of multi-class image classification problem where we calculate the class probabilities by using the softmax function and then calculating the negative log loss over them. This works well for the single label classification problems but it is not used for multi-label classification problems where we have multiple labels of the same image and we try to maximize the scores of each of the labels. Instead another loss function called Binary Cross Entropy function (Eq \ref{eq:binary-cross-entropy})
\begin{equation}
\label{eq:binary-cross-entropy}
    \ell(y, p) = -\sum_{i=1}^{C}y_i\log p_i - \sum_{i=1}^{C}(1-y_i) \log (1-p_i) 
\end{equation}
In the equation \ref{eq:binary-cross-entropy}, $y$ is the true label, $p$ are the scores between $0$ and $1$, and $C$ is the total number of classes. In our problem, since we want to minimize the loss between many similar images, we extend the binary cross entropy term to calculate the loss between all the image representations. The scores $p(x_i, \tilde{x}_j)$ can be replaced the cosine similarity between the $i^{th}$ and $j^{th}$ representations and the labels $y(x_i, \tilde{x}_j)$ can be computed using the threshold values of the similarity score as we did in Section \ref{sec:ccsl}. More specifically, let $\Bar{q}_{z_{\theta}}(x_i)$ and $\Bar{z}'_{\xi}(x_j)$ be the image representations of images $x_i$ and $x_j$ from the online and target networks respectively. Then the Cross-Sigmoid Similarity Loss(CSSL) can be calculated using the equation \ref{eq:cssl-per-example} 
\begin{equation}
\label{eq:cssl-per-example} 
\begin{split}
    \ell(x_i,x_j) &= \ell(\Bar{q}_{z_{\theta}}(x_i), \Bar{z}'_{\xi}(x_j)) \\
    \ell(\Bar{q}_{z_{\theta}}(x_i), \Bar{z}'_{\xi}(x_j)) &=  -\mathbbm{1}\Big[\langle \Bar{q}_{z_{\theta}}(x_i), \Bar{z}'_{\xi}(x_j)\rangle \geq \theta_p \Big] \log p \Big( \Bar{q}_{z_{\theta}}(x_i), \Bar{z}'_{\xi}(x_j) \Big)\\
    & \hspace{20mm} -\mathbbm{1}\Big[\langle \Bar{q}_{z_{\theta}}(x_i), \Bar{z}'_{\xi}(x_j)\rangle \leq \theta_n \Big] \log  \Big( 1 -  p(\Bar{q}_{z_{\theta}}(x_i), \Bar{z}'_{\xi}(x_j)) \Big)
\end{split}
\end{equation}
In the equation \ref{eq:cssl-per-example}, if the similarity of the representations are high then they will get even more closer because of the first term, and if the similarity is very less, then they will get further apart due to the second term. Using the described loss term, the loss function for the input image $x_i$ can be written as 
\begin{equation}
\label{eq:cssl-single}
    \mathcal{L}_{CSSL}(x_i) = -\log \Big(p(\Bar{q}_{z_{\theta}}(x_i), \Bar{z}'_{\xi}(x_i))\Big) - \lambda \sum_{i\neq j}\ell (x_i, x_j)
\end{equation}
In the equation \ref{eq:cssl-single}, the first term is the loss between the representations of the online and target networks for the same image, and the second term is the loss term of the online representation of $x_i$ with the target representations of $x_j$,  $\forall j$ such that $j\neq i$, in the input mini-batch. The weighting parameter $\lambda$ is used to avoid the representational collapse. Note that in this CSSL loss we have also taken the loss from the dissimilar images which was not the case in the CCSL loss function. Finally to keep the symmetry between the online and target networks, we interchange their inputs and calculate the loss $\tilde{\mathcal{L}}_{CSSL}(x_i)$ and the final loss is given by the equation \ref{eq:cssl-final}
\begin{equation}
\label{eq:cssl-final} 
    \mathcal{L}_{CSSL}^{BYOL}(x_i) = \mathcal{L}_{CSSL}(x_i) + \tilde{L}_{CSSL}(x_i) 
\end{equation}
The loss across the mini-batch is the average loss over all the images. 

% BYOL is said to be collapsed when it produces either completely positive or negative results. There has been a hypothesis that Batch Normalisation (BN) is essential for eliminating collapse in BYOL. \cite{NoStatistics} have proposed a batch independent normalisation scheme that replaces BN and produces results comparable to BN version. In this project, we will reproduce their method to analyse this finding. 

\begin{figure}[t]
    \begin{minipage}{0.3\linewidth}
        %\centering  % redundant
        \includegraphics[width=\textwidth]{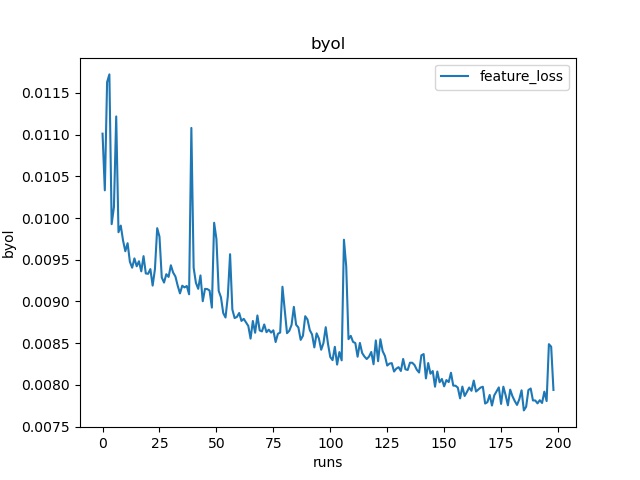}
        \caption{Feature Loss across training epochs of the Vanilla L2 Loss Function of BYOL}
        \label{fig:vanilla_byol_loss}
    \end{minipage}%
    \hfill% not: "\hspace{0.5cm}"
    \begin{minipage}{0.3\linewidth}
        %\centering  % redundant
        \includegraphics[width=\textwidth]{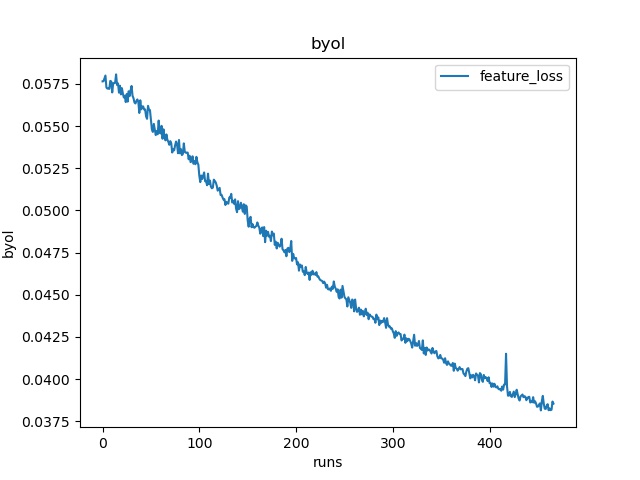}
        \caption{Feature Loss across training epochs of the CCSL Loss Function of BYOL}
        \label{fig:ccsl_loss}
    \end{minipage}%
    \hfill% not: "\hspace{0.5cm}"
    \begin{minipage}{0.3\linewidth}
        %\centering  % redundant
        \includegraphics[width=\textwidth]{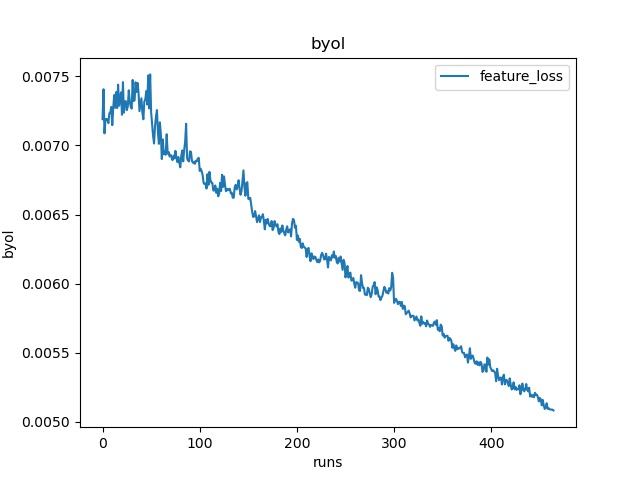}
        \caption{Feature Loss across training epochs of the CSSL Loss Function of BYOL}
        \label{fig:cssl_loss}
    \end{minipage}
\end{figure}
\begin{figure}[t]
    \begin{minipage}{0.3\linewidth}
        %\centering  % redundant
        \includegraphics[width=\textwidth]{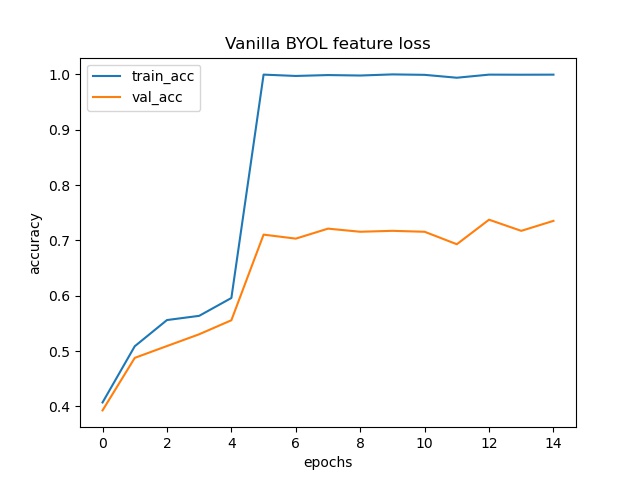}
        \caption{Linear Evaluation Accuracy of Vanilla BYOL.}
        \label{fig:vanilla_byol_acc}
    \end{minipage}%
    \hfill% not: "\hspace{0.5cm}"
    \begin{minipage}{0.3\linewidth}
        %\centering  % redundant
        \includegraphics[width=\textwidth]{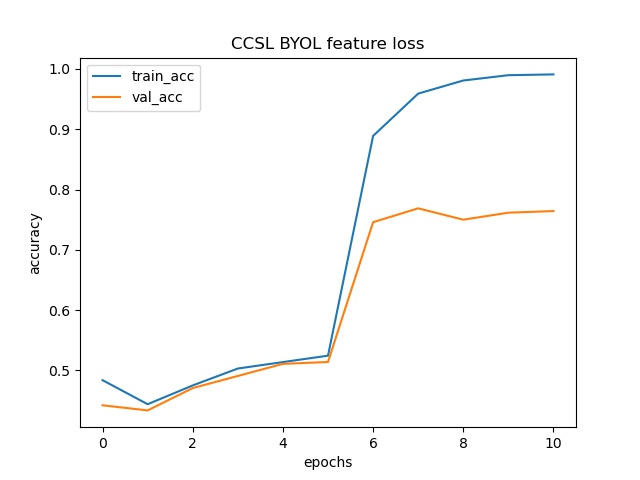}
        \caption{Linear Evaluation Accuracy of CCSL BYOL.}
        \label{fig:ccsl_acc}
    \end{minipage}%
    \hfill% not: "\hspace{0.5cm}"
    \begin{minipage}{0.3\linewidth}
        %\centering  % redundant
        \includegraphics[width=\textwidth]{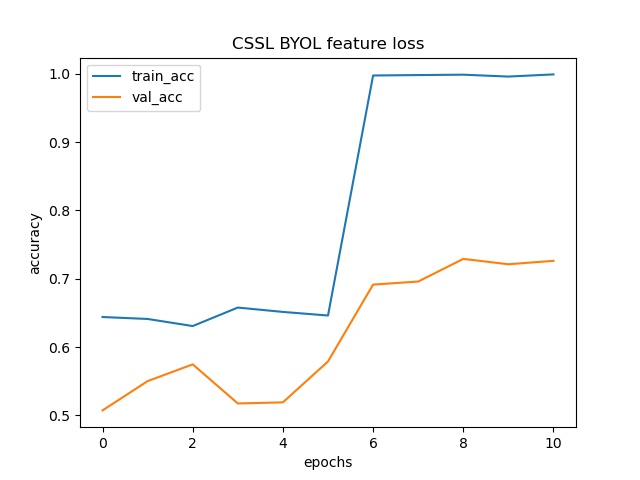}
        \caption{Linear Evaluation Accuracy of CSSL BYOL.}
        \label{fig:cssl_acc}
    \end{minipage}
\end{figure}

\section{Experiments and Results}

% \begin{table}[b]
%     \centering
%     \begin{tabularx}{\textwidth}{|>{\centering\arraybackslash}X|>{\centering\arraybackslash}X|>{\centering\arraybackslash}X|}
%     \hline
        
%         \textbf{Configuration} & \textbf{Accuracy}  \\
%     \hline
%     Without BN & 52.46\%  \\
%     \hline
%     Without BN + WS & 53.12\% \\
%     \hline
%     LN  & 60.03\%\\
%     \hline
%     LN + WS & \textbf{69.65}\%\\
%     \hline
%     GN  & 62.96\%\\
%     \hline
%     GN + WS & \textbf{70.11}\%\\
%     \hline
%     Vanilla BYOL & 71.04\%\\
%     \hline
%     CCSL BYOL & \textbf{76.87}\%\\
%     \hline
%     CSSL BYOL & 7\textbf{72.92}\%\\
%     \hline
%     \end{tabularx}
%     \caption{Accuracies obtained on different configurations. Bolded figures confirm expected high results.}
%     \label{tab:result_tab}
% \end{table}

\begin{table}
	\begin{minipage}{0.5\linewidth}
		\label{table:accuracies}
		\centering
		\begin{tabular}{lrr}
			\toprule
			\textbf{Configuration}  & \textbf{Accuracy} \\
			\midrule
			Without BN       & 52.46\%  \\      
			Without BN + WS  & 53.12\%  \\    
			LN               & 60.03\% \\    
			LN + WS          & \textbf{69.65\%} \\    
			GN               & 62.96\%\\    
			GN + WS          & \textbf{70.11}\%  \\    
			Vanilla BYOL     & 71.04\%  \\    
			CCSL BYOL        & \textbf{76.87}\%  \\    
			CSSL BYOL        & \textbf{72.92}\%  \\    
			\bottomrule
		\end{tabular}
		\caption{Accuracies obtained on STL10 dataset with different model configurations. Bolded figures confirm expected high results.}
	   \label{table:accuracies}
	\end{minipage}\hfill
	\begin{minipage}{0.60\linewidth}
		\centering
		\includegraphics[width=70mm]{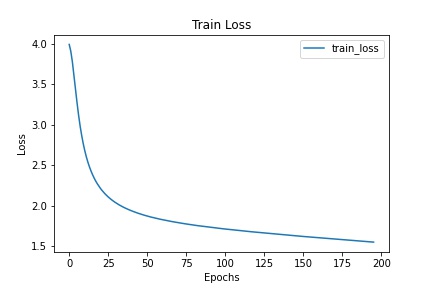}
		\captionof{figure}{Smoothened Train Loss in \\ LN+WS model}
		\label{tab:result_tab}
	\end{minipage}
\end{table}

\subsection{Dataset}
We will use STL10\cite{stl10dataset} dataset for all our experiments. The dataset consists of 100,000 unlabeled RGB images with $96\times 96$ pixels. There is also small labeled dataset with 500 training images divided into 10 classes and 800 test images per class. The unlabeled partition of the dataset is used to train the BYOL network and the labeled partition is used to evaluate the learned representations using the Linear Evaluation protocol described Section \ref{sec:linear_eval}

\subsection{Network Architecture}
\label{sec:network_arch}
We have used ResNet-18 \cite{resnet} as online network $f_{\theta}$ and target network $f_{\xi}$. The output representations of the ResNet-18 before the fully connected layer is of size 512. The projector networks for both the online and target networks are 2-layer neural networks with Batch Normalization and ReLU activation function. The input size, hidden size, and the output sizes are all equal to 512. The predictor head of the online network is 2-layer network with same configurations as the projector networks. We have used SGD with momentum of 0.9, learning rate of 0.008, and batch size of 512 to train the network. To reduce the over fitting and regularizing the network we have also inserted Dropout layers between the Bottleneck modules of the ResNet-18 layers with the probability of dropping the weights as $0.1$.
\subsection{Linear Evaluation Protocol}
% \subsubsection{ResNet-18 with Dropout}

\label{sec:linear_eval}
To evaluate the quality of the representations learned by the online network of BYOL, we take the online network and freeze its weights. Then a randomly initialized fully connected layer is trained on top of the online network using the labeled dataset. After some training runs of BYOL, we train the linear image classifier using random fully connected layer to convergence. From Figure \ref{fig:vanilla_byol_acc}, we can see that as the training of BYOL progresses, the the learned representations are able to achieve consistently higher accuracies. 

\begin{figure}[t]
    \begin{minipage}{0.3\linewidth}
        %\centering  % redundant
        \includegraphics[width=\textwidth]{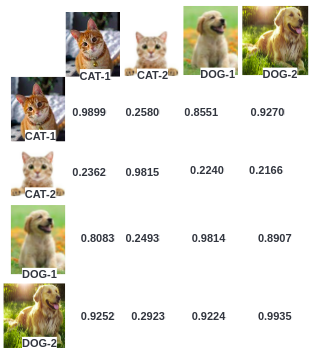}
        \caption{The similarity scores between different images using the features from the Vanilla BYOL representations.}
        \label{fig:vanilla_byol_representations}
    \end{minipage}%
    \hfill% not: "\hspace{0.5cm}"
    \begin{minipage}{0.3\linewidth}
        %\centering  % redundant
        \includegraphics[width=\textwidth]{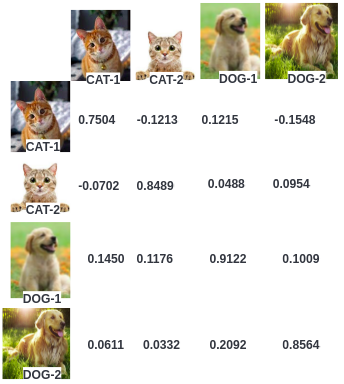}
        \caption{The similarity scores between different images using the features from the CCSL BYOL representations.}
        \label{fig:ccsl_representations}
    \end{minipage}%
    \hfill% not: "\hspace{0.5cm}"
    \begin{minipage}{0.3\linewidth}
        %\centering  % redundant
        \includegraphics[width=\textwidth]{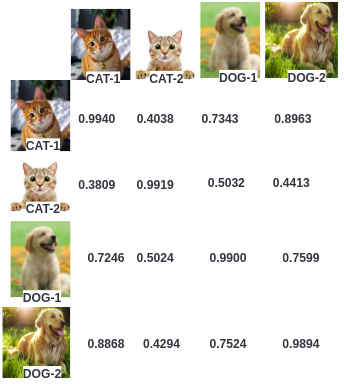}
        \caption{The similarity scores between different images using the features from the CSSL BYOL representations.}
        \label{fig:cssl_representations}
    \end{minipage}
\end{figure}

\subsection{Results}
\label{sec:results}
\subsubsection{Batch Statistics Experiments}
We begin by investigating the importance of BN in BYOL framework. For that purpose, we removed all the BN layers in both encoder and target network. We found that the performance is nearly random. However, keeping BN in the only the encoder results in boost in performance as compared to the without-BN version. We were keen to find out the whether is it the batch statistics which induces contrastive learning implicitly within the Vanilla BYOL framework. So, we replaced BN with Layer Normalisation (LN) \cite{ln} and Group Normalisation (GN) \cite{gn} which do not require batch statistics. From Table \ref{table:accuracies}, we see that performance of BYOL with LN or GN has boosted up after adding WS in convolutional layers followed by GN or LN. WS has also helped us reducing the training time drastically. Comparing Fig. 2 and Fig. 8, we observe that the spikes in Figure 2 in case of Vanilla BYOL have vanished and loss landscape has become smooth. We also infer from Table \ref{table:accuracies} that GN+WS and LN+WS versions are performing comparable to the vanilla BYOL. This shows that BYOL is able to learn non-collapsed representation without BN.  

\subsubsection{Self-Labeling Refinement Results}
In this section we describe the results of the experiments conducted in this work and their analysis. Figures \ref{fig:vanilla_byol_loss}, \ref{fig:ccsl_loss}, and \ref{fig:cssl_loss} shows the representation loss and figures \ref{fig:vanilla_byol_acc}, \ref{fig:ccsl_acc}, and \ref{fig:cssl_acc} shows the results for the linear evaluation protocol for the vanilla BYOL, CCSL BYOL and CSSL BYOL respectively. From the Table \ref{table:accuracies}, we can see that the representations from the CCSL BYOL gives the best accuracy on the STL10 validation dataset and CSSL BYOL gives similar results to the Vanilla BYOL setting. For qualitative analysis of the learned image features, we take the output representations of the online network and the target network for various images and computed the similarity between them using the cosine similarity or the dot product of the normalized representations. From Figure \ref{fig:vanilla_byol_representations}, we can see that the similarity between the image representations is highest along the diagonal, which can be expected because the online network and target network is giving the representations for two different views of the same image. However, in the same figure \ref{fig:vanilla_byol_representations}, we can see the high similarity scores in the first row, between cat-1, and dog-1 which is 0.8551 but low score between cat-1 and cat-2 which is 0.2580. Intuitively, the similarity scores between the images of cat-1 and cat-2 should be higher than the similarity score between cat-1 and dog-1. But in the case of vanilla BYOL, we can see that there is a discrepancy. More examples of this discrepancy can be seen in the same figure between the dog-1 and cat-1 in the $3^{rd}$ row and between dog-2 and cat-1 in the $4^{th}$ row. 

The proposed CCSL and CSSL loss functions try to mitigate this discrepancy and improve the similarity scores between the semantically similar images. After training the BYOL with CCSL loss function, we produced the similarity scores. The similarity scores between the images are presented in figure \ref{fig:ccsl_representations}. The similarity scores along the diagonal is highest but 'regularized' because the scores are less than corresponding scores we got with vanilla BYOL representations. From the $1^{st}$ row of figure \ref{fig:ccsl_representations}, we can see that score between cat-1 and dog-1 is low as compared to the similarity scores that we got with the vanilla BYOL. Similarly, in figure \ref{fig:cssl_representations}, the scores in the $1^{st}$ row between the cat-1 and dog-1 decreased slightly as compared to the scores with vanilla BYOL. Also note that while some scores are showing the expected behavior, some of the scores are erratic. For instance, in figure \ref{fig:ccsl_representations}, the score between cat-1 and cat-2 in the $1^{st}$ row is negative when it should be positive. We attribute this erratic behavior to the size of the unlabeled dataset. If we increase the amount of the unlabeled images, we might see expected scores between the semantically similar and dissimilar images.

\subsection{Conclusion}
In our project, we have analysed the representation learning in BYOL. We conducted experiments to confirm that normalisation schemes other than BN  like GN and LN perform comparably when incorporated with weight standardization. We were also show that the performance of BYOL with explicit constrastive terms in the loss expression. The CSSL loss function gives better performance than vanilla BYOL and CCSL performs comparably. To this end, we conclude that BYOL benefits with the proposed loss functions. For the future work, we would want to push the performance of BYOL with the proposed loss functions on the ImageNet \cite{deng2009imagenet} dataset which is a standard benchmark. We would work to rectify the failure cases in Fig. 10 and Fig. 11 by fine-tuning the neural network using the different values of $\lambda$.

% \subsection{Analysing Batch statistics for BYOL}
% We will try to analyse how BN is able to avoid collapse results in BYOL by removing BN and comparing the performance on STL-10 dataset. 
% We will perform another set of experiments to confirm that BYOL can produce non-collapse results event without BN. We will apply different weight standardization schemes to convolutional and linear parameters, and then replace BN or LN with Group Normalisation.

% We will compare results from the encoders trained using the methods described in Section \ref{sec:methods} and the encoder with pre-trained ImageNet \cite{deng2009imagenet} weights.

% \subsection*{Collaboration}
% There is no collaboration of this project with any other class.
% \subsection*{Work Distribution}
% \begin{itemize}
%     \item Siddhant Garg: Implementation and training of BYOL Architecture from scratch (Sec.\ref{sec:byol}), Proposed Self-Labeling Refinement (Sec.\ref{sec:self_labeling}),  and implemented and trained BYOL with CCSL loss function (Sec.\ref{sec:ccsl}) and CSSL loss function (Sec.\ref{sec:cssl}). Implementation of Linear Evaluation Protocol (Sec. \ref{sec:linear_eval})
%     \item Dhruval Jain: 
% \end{itemize}
\printbibliography
\end{document}